\documentclass[a4paper]{article}\usepackage{pb-diagram,lamsarrow,pb-lams}
\usepackage{amsmath, amsthm, amsfonts, microtype,siunitx,booktabs,graphicx,amscd,verbatim,makeidx,bbm,emptypage,titleps, array}
\usepackage[pdf]{xy}
\usepackage{tabu,tabularx,caption}
\usepackage{fancyvrb}
\usepackage{tikz}
\usepackage{fancyhdr}

\theoremstyle{definition}

\VerbatimFootnotes

\begin{document}

\title{Named entity recognition using conditional random fields with non-local relational constraints}
\author{F.\ M.\ Cecchini\footnote{Universit\`a degli studi di Milano, flavio.cecchini@studenti.unimi.it}, E.\ Fersini\footnote{University of Milano--Bicocca, fersiniel@disco.unimib.it}}
\date{}
\maketitle

\begin{center}\section*{Abstract}\end{center}
\scriptsize We begin by introducing the Computer Science branch of Natural Language Processing, then narrowing the attention on its subbranch of Information Extraction and particularly on Named Entity Recognition, discussing briefly its main methodological approaches. It follows an introduction to state-of-the-art Conditional Random Fields under the form of linear chains. Subsequently, the idea of constrained inference as a way to model long-distance relationships in a text is presented, based on an Integer Linear Programming representation of the problem. Adding such relationships to the problem as automatically inferred logical formulas, translatable into linear conditions, we propose to solve the resulting more complex problem with the aid of Lagrangian relaxation, of which some technical details are explained. Lastly, we give some experimental results.\normalsize

\bigskip

\section{Introduction}

Since one of the earliest systematic forms of writing we know was used by Sumerians at least 5000 years ago for book-keeping, uncountable ways of fixing human words and thoughts on a durable support have been created, developed and spread nearly everywhere in the world, becoming an unavoidable pillar of modern complex societies. In a way, the "computer revolution" of the last decades, surely giving an unprecedented boost to visual and auditive information, has at the same time exasperated the magnitude of written text we can get our hands on: We are now used to read in newspapers or in the internet, replicated in many forms, what we would have once only heard from a friend or a town-crier. Since then communication has been changing and computer hardware has been rapidly gaining in power and flexibility, one of the most obvious paths for research to take in the field of Computer Science has been so-called Natural Language Processing.

\bigskip

Natural Language Processing (NLP for short) is the vast branch of Computer Science trying to find out valid methods to let humans and computers interact by means of natural languages. By "natural languages" we intend the system we use every day to articulate sounds in precise grammatical constructs so as to convey meanings to our fellow humans: in other words, when we speak, we are using a natural language. We call them natural in contrast to artificial or formal ones developed by software engineers to write their programs. In fact, humans have always needed some sort of language or code to impart instructions to machines, but it is far easier to create one for such purposes than to adapt an already existing natural one: Artificial languages are very limited, but completely regular and alterable at will. Natural languages instead are not arbitrary and have a life of their own: their structure can be very complex and present many variations, due to their millenary constant use. This complexity goes hand in hand with the infinite possibilities of expression that every natural language enables and it is what makes them so difficult to manipulate mechanically.

\bigskip

NLP is a clearly interdisciplinary science involving Linguistics, Computer Science, Mathematics and reaching even into Psychology and Science of Education with its applications\footnote{More detailed general introductions to NLP, its subbranches and its basic techniques mentioned in this section can be found in \cite{Fer} or \cite{Jur}.}. Mutual constructive influences have not been unknown, especially between the first two areas (e.\ g. with Chomsky). The part of NLP more practically engaged with its software implementation and connected problematics is called Computational Linguistics.\\
Starting from the utopian aim of a ``fully automatic high quality translation'', many other subbranches of NLP have stepped up and gained in importance, like Machine Aided Translation (\cite{Sl}), Man-Machine Interaction (\cite{Son}), Text Understanding (\cite{Fen}), Text and Report Generation (\cite{McK}), and so on, reaching up to the domain of Artificial Intelligence. Attention has been so widened from processing of written sources and written interaction to verbal interaction and even to communication enhanced by non-verbal components. However, one could distinguish since the beginnings two broad categories of tasks. The one at a more theoretical level is concerned with creating models of human comprehension and production of the language and has been very much influenced by Generative linguistics initially (\cite{Cho}); the second and more practical one tries to isolate words from texts, reorder and count them, so as to establish databases for successive statistical treatment (\cite{Bu1}, \cite{Bu2}). In both cases, the notion of syntax and syntactical analysis plays a major role.

\bigskip

Syntactic analysis, or parsing, is a cornerstone of NLP. Its goal is to decompose a sentence into its parts of speech, that is, its fundamental syntactical constituents, each having its own grammatical role, at the same time specifying the relations intercurring between them. The whole picture gives the structure of the sentence, from which further investigation or manipulation can be pursued. Many strategies have been developed for syntactical analysis, and they often involve one of two major strategies, either bottom-up or top-down\footnote{Again, \cite{Fer} and \cite{Jur} for a more detailed discussion thereof.}. The theory of syntactic (and lexical, but even morphological) analysis has strong ties with Automata Theory and therefore has been thoroughly developed, but still holds great margins of improvement due to the intrinsic complexity and ambiguity of natural languages, and it is often only a good approximation. Nonetheless, by parsing we can decompose a sentence, or a whole text, in bits of processed information that can later be used by a machine.

\section{Information Extraction and NER}

Completely mechanically "decoding" a sentence (and even more so a text) as described before, however, can present until now overwhelming difficulties or ambiguities. In this sense, it could be reasonable to moderate the most theoretical requests and to restrict the attention only to a portion of the meaning conveyed by the text: this is the case of Information Extraction (IE), a branch of NLP devoted, as the name implies, to extrapolating just the relevant pieces of information for a specific inquiry.

\bigskip

Let's first say that IE approaches have proven very successful until now, but can incur into two major hindrances (see \cite{Zh} for an introduction). The first one is that an IE program often bears an inherent dependency from the specific and limited domain it was thought for; this could lead to problems of portability. For example, an IE method conceived for a database about ships could perform not so well for one about cats. Due to the fact that in practical applications IE instruments have to be implemented for new domains in a reasonable amount of time and that it would be very time-consuming to always start anew, IE systems should strive to be as independent from human readjustments and intervention as possible. \newline
The other major issue is the quality and lack of training data. Prior to being applied to free text, supervised and semi-supervised (see below) IE programs have to be trained (i.\ e. have to set their parameters for future evaluations) on a training corpus where key elements are annotated beforehand; details about the training can widely differ, but annotation on large corpora is generally a tedious and expensive job. As a consequence, training sets tend to be sparse and relatively small and IE methods for a certain domain have to generalize the most and correctly from a limited amount of starting material. This of course doesn't apply to unsupervised learning methods.\\
Both problems plot into a third one not directly connected with computational implementation, that is, the objective evaluation of the effectiveness of an IE program: It is difficult to find criteria valid for every domain, and the variable quality of training data can have a sensible impact on performances (see about this topic \cite{eva}).

\bigskip

Putting aside these fundamental issues, more than taking into account e.\ g. offline newspapers or linguistic corpora, Information Extraction tools are becoming increasingly useful due to the ever-growing mass of written data uploaded on the web and their instant accessibility; the need to manage them and to store the information contained therein in structured files has led to the development of different possible tasks, which we could arrange hierarchically (again \cite{Zh}): Named Entity Recognition (NER), Entity Relation Extraction and Event Extraction. From the particular to the whole picture, their respective aims are: to spot key sequences of words in a text and identify them correctly; to establish bonds and correlations between them; and finally, to be able to infer complete events based on previously found clues, so that it could be possible for us to interrogate a database about them. Even if the difficulties faced by IE could seem not so imposing at first, they appear discouraging if we just take a moment to think about the different ways we use every day to refer to people or facts: Matilde di Canossa could be just Matilde, or Mathilde, or Grancontessa Matilde, Grancontessa or simply "she", and every term could be perceived as distinct from the other ones by a machine, while we know they coincide. We'll now quickly detail some strategies regarding this  problem found in NER.

\subsection{Named Entity Recognition}

The objective of NER is to identify single, elementary unities in a text which could very well appear in a free and hence not pre-processed form, and to assign them pre-defined categories and templates; names of persons or places are maybe the first examples coming to mind. In fact, the notion of ``named entity'' was standardized during MUC-7, the seventh Message Understanding Conference, an IE competition financed by US government which was held from 1987 to 1997\footnote{MUC-7: \verb|http://www-nlpir.nist.gov/related_projects/muc/proceedings/ne_task.html|. }. The crucial point is that an algorithm has to be able to recognize unknown parts of text based on some kind of induction rules. Since this task involves to some extent an analysis of texts not dissimilar in its basics from parsing, first approaches were based on fixed cascading rules manually implemented by human experts, which could be regarded as some kind of axioms (for this and the following discussion, see \cite{survey}). The effort going into the compilation of these rules, which could sensibly vary from one domain of application to another as mentioned earlier, is remarkable and thus has pushed researchers to find a way to automatize it. Various NLP and Machine Learning techniques have proven crucial in creating systems which can inductively derive ordered sets of rules (thought of initially as Boolean conditions) from previously labelled texts (\cite{rau} is a pioneering work in this sense), but there are other approaches too.

\bigskip

We can distinguish between three main kinds of automatic learning: supervised, semi-supervised and unsupervised (again \cite{survey}).\\
Supervised learning is still the dominant technique and comprises many of the most studied methods, many of which of statistical imprint (see below). The founding idea is to build a sufficiently large dataset or corpus who has to be manually tagged; this human intervention aside, the learning method then automatically assimilates needed information and adjusts its parameters and adds disambiguation rules using its implemented features. As mentioned before, the main issue about supervised learning is the need for extremely big corpora, whose availability is scarce and whose cost is prohibitive.\\
Semi-supervised learning advances are relatively more recent; one of the first studies about them can be found in \cite{brin}, and more recently in \cite{CV}, \cite{pasca} or \cite{HG}. Such a system, whose goal is trying to extrapolate rules for a given entity, has first to be seeded, by being e.\ g. handed by the user a list of names relating to the chosen entity (and possibly a small set of lexical or syntactical rules or similar). It then begins to find those names and the contexts around them and successively goes on to catch other names of the same type using its new-found context clues.\\
Unsupervised learning aims instead at clustering, that is at finding different types of entities that share common patterns. In contrast with supervised and semi-supervised techniques, there is no starting point to which the found results can be compared. In fact, unsupervised learning detaches from a rule-inductive approach in the sense that it doesn't actually defines rules, but sets, establishing for each element if it fits in a set or another. As an example of such methods, in \cite{sese} the central observation is that named entities often appear simultaneously in different newspaper articles, whereas common names don't. Other examples are found in \cite{AM} or \cite{Etz}.

\bigskip

Still belonging to the realm of rule-inductive (and mostly supervised) methods, but taking a different path from pre-defined hand-written rules, is the statistical approach, which has been more and more successfully forwarded in the last years. In this perspective, a NER algorithm is a process which we require to make decisions by assigning a sequence of labels to a first sequence of observations, viewed as possibily joint random variables of which the algorithm has to guess the value. Every time, the algorithm decides based on its training and on what it has observed precedently. There are two different ways to do this: one takes into consideration and labels every token of the input (e.\ g. \cite{TC}), and the other one instead only segments of the input, not necessarily coinciding with tokens (e.~g.~\cite{SC}).\\
The most widespread statistical methods are the Hidden Markov model (HMM), the Maximum Entropy model and Conditional Random Fields.

\bigskip

Heuristically, a sentence could be imagined as a net where every knot represents a possible value of an observed variable: to somehow label the sentence corresponds then to the choice of a path from one to the other end of the net. This is the intuition lying behind the concepts of graph and Markov chains applied to syntactic analysis, all finding use in NER. Specifically, in the HMM we try to find for every token in the sequence of observations its true, hidden underlying "state" by guessing from its superficial value (an introduction can be found in \cite{MS}, chapter 9). We can assume then that the status of every token depends only from the the previous one with a given probability, and it is here that the Markov hypothesis kicks in. Starting from token one and going on, we can represent for every step its probable states as nodes in a graph, which are in turn connected to the nodes of the following step. Every connection between two nodes has its probability; we could see it as the weight of that connection. Then, we could trace all possible paths in the graph from the first to the last token. However, we are interested in the most probable one, that is, the path whose weight is heaviest. This can be done via the Viterbi algorithm.\\
The Maximum Entropy model (MEM) tries to extract rules and constraints regarding the possible values of the label random variables, and then readjusts their probabilities accordingly in such a way that maximum entropy is reached; in other words, probabilities must be distributed as evenly as possible respecting the constraints. It can proved that such a distribution always exists and is unique and an algorithm converging to it can be constructed (\cite{pietre}).\\
However, now it is Conditional Random Fields (CRF) that are regarded as state of art. They take steps from a Markovian point of view and are detailed in the next section.

\section{Conditional Random Fields}

A Conditional Random Field is defined (see \cite{crf} for the following discussion) starting from an indirected graph $G=(V,E)$ with edges $E$ and whose vertices $V$ index the components of the random variable $Y=(Y_1,\ldots,Y_n)$ over the label sequence to be assigned; if we call $X=(X_1,\ldots,X_n)$ the random variable over the observed sequence to be labelled, then $(X,Y)$ is a conditional random field if the random variables $Y_i$ conditioned to $X$ satisfy the Markov property with respect to their neighbours in the graph. In other words, a CRF is a random field globally conditioned on the observation $X$. The simplest form of a graph we can think of is a linear chain, which is nevertheless very useful to model a sequence of observations making up a sentence; sentences do possess a somewhat linear nature, as they are sequences of words. Since a linear chain is also a tree, it is possible to express the joint distribution $p(y|x)$ of labels $y$ given $x$ in a precise form, thanks to the fundamental theorem of random fields (\cite{HC}):
\begin{equation}\label{campcasuali}p(y|x)=\frac{1}{Z(x)}\exp(\sum_{e\in E,k}\lambda_kf_k(e,y_{|e},x)+\sum_{v\in V,k}\mu_kg_k(v,y_{|e},x)).\end{equation} Here $y_{|e}$ and $y_{|v}$ are the components of $y$ respectively associated with an edge or a vertex of $G$, and $Z(x)$ is the observation-dependent normalization factor. Probability \eqref{campcasuali} is obtained noting that the cliques of a linear chain are its vertices and its couples of vertices connected by an edge.\newline
The factors $f_k$ and $g_k$ are feature functions: They are fixed and give us a measure of the distinctive traits every label and its context possess. In case of Boolean functions, for example, they tell us about the presence or absence of such traits. Each feature has an associated weight $\lambda_k$ or $\mu_k$ as a free parameter which refers to the importance of that feature in determining the probability of $y$ given $x$. Features so introduced are a logical detail in a statistical structure: They could be expressed as logical formulas, true when a combination of simultaneous factors occurs. As an example, a feature will have the value 1 if an observed word is labelled as a proper name and is followed by another proper name label. Features come in two types: transition features, corresponding to an edge connecting two labels, and state features, corresponding to single vertices.

\bigskip

CRFs are based on a Markovian representation of events, like the HMM described before. Now, the main difference between HMM and CFRs is that the former can be represented by mean of a directed graph, whereas the latter is indirected.
\\
CRFs in the form of linear chains still retain a local nature though, and the problem persists of modelling long-distance relationships. In order to include them in the model to some degree and to enhance the power of CRFs, many solutions have been suggested and two main paths have emerged: either one could try to relax the Markov assumptions at the base of the model (\cite{SC}, \cite{Gal}), or extra information could be added in the form of logical constraints during the inference phase (\cite{K}, \cite{RY}, \cite{Ch1}, \cite{Ch2}). The second solution has the lesser impact on computational complexity. Indeed, one could also consider non linear chain CRFs to include some sort of non-local relationships, but then the determination of cliques in the resulting graph (which is by the way an NP-complete problem), needed to compute probability \eqref{campcasuali}, and the training of the parameters corresponding to the features would easily reach incredibly high levels of intricacy.

\subsection{Constrained inference}

The introduction of further knowledge in the CRF model could help correcting local errors in the predictions by providing global relations that should be satisfied. This can be done by translating the labelling problem into an Integer Linear Programming (ILP) one. In fact, we again return to the linear chain representation of label random variables. From the probability in equation \eqref{campcasuali}, we can define (see again \cite{crf}) for every step $t$ in this chain a matrix $M_t(x)$, depending from the observations, whose entries $M_t(y,y'|x)$ (where $y'$ and $y$ are possible labels) are essentially proportional to the probability of passing from label $y$ to label $y'$ at time $t$. We see them as weights of edges in the graph representing every possible assignment of labels to observations. Here is an example corresponding to the case where the observation sequence has three elements (``words'') and the possible labels are three. Two special starting and ending nodes, respectively at times 0 and $n+1$, where $n$ is the length of the sequence, have been introduced.

\bigskip

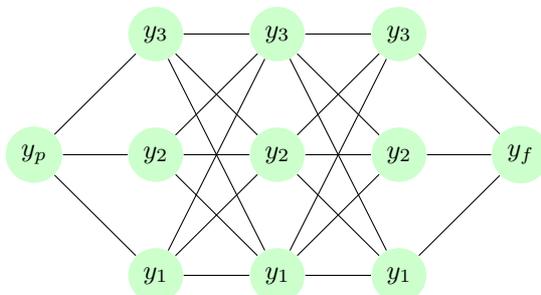
\begin{center}
\begin{tikzpicture}
  [scale=.8,auto=left,every node/.style={circle,fill=green!20}]
  \node (n1) at (1,3)  {$y_p$};
  \node (n2) at (3,1)  {$y_1$};
  \node (n3) at (3,3)  {$y_2$};
  \node (n4) at (3,5)  {$y_3$};
  \node (n5) at (5,1)  {$y_1$};
  \node (n6) at (5,3)  {$y_2$};
  \node (n7) at (5,5)  {$y_3$};
  \node (n8) at (7,1)  {$y_1$};
  \node (n9) at (7,3)  {$y_2$};
  \node (n10) at (7,5) {$y_3$};
  \node (n11) at (9,3) {$y_f$};

  \foreach \from/\to in {n1/n2,n1/n3,n1/n4,n2/n5,n2/n6,n2/n7,n3/n5,n3/n6,n3/n7,n4/n5,n4/n6,n4/n7,n5/n8,n5/n9,n5/n10,n6/n8,n6/n9,n6/n10,n7/n8,n7/n9,n7/n10,n8/n11,n9/n11,n10/n11}
    \draw (\from) -- (\to);

\end{tikzpicture}\end{center}

\bigskip

The upper left edge will have e.\ g. weight $M_1(y_3,y_3|x)$ here.\\
We have then come back to the problem of finding the heaviest, that is the most probable path on the graph. The problem can be expressed in the language of ILP: Namely, if we take the logarithmic value of each $M_t(y,y'|x)$ as entries of an $(n-1)m^2+2m$-dimensional ($n$ be the length of the sequence and $m$ the number of possible labels or states) vector $M$ and define the Boolean variable \[\begin{split}e_{t,yy'}&=1\quad\text{if the edge from $y$ to $y'$ at time $t$ is in the most probable path}\\ &=0\quad\text{otherwise}\end{split}\] forming a vector $e$ with the same dimension of $M$, we can proceed to formulate the problem as follows.\\\\

\begin{equation}\begin{split}\label{ilp} &\max Z(e)=M^t\cdot e\\ \text{subject to}\quad&\sum_{y\in\tilde{\mathcal{Y}}}e_{t-1,y\hat{y}}-\sum_{y\in\tilde{\mathcal{Y}}}e_{t,\hat{y}y}=0\quad\forall t\quad\text{t.c.}\quad1\leq t \leq n,\quad\forall \hat{y}\in\mathcal{Y}\\ &\sum_{y\in\mathcal{Y}}e_{0,y_py}=1,\quad\sum_{y\in\mathcal{Y}}e_{n,yy_f}=1.\end{split}\end{equation}

\bigskip

Here $\mathcal{Y}$ is the set of all possible labels and $\tilde{\mathcal{Y}}=\mathcal{Y}\mathcal\cup\{y_p,y_f\}$. The first set of constraints defines the path: at every time $t$ only one node corresponding to a status $y\in\mathcal{Y}$ can be visited, and at every node exactly one edge of the path enters and another one exits. The last two constraints assure us that only exactly one edge in the path exits from the starting point $y_p$ and, conversely, exactly one arrives at $y_f$.\\
Now, if we were to represent other constraints or especially long-distance relationships between the components of a sentence, which is not possible with Viterbi algorithm (whose nature is only local), we could add them as further linear equalities or inequalities to the main problem \eqref{ilp}. Hence new questions arise: how to define them and what does it mean to satisfy them? The approach found e.\ g. in \cite{Zu} requires again human intervention and that domain experts set these new rules by hand. Further, once fixed, the new constraints must be satisfied. The time-consuming activity of defining them aside, there is a complicating issue with them being every time consistent with new training or test data: Possible impracticable solutions could originate otherwise. As a conclusion, there is a need for greater flexibility and for a system that is able to take into account for errors.

\bigskip

If we consider sentences, their syntactic structures are patterns, which in turn contain repeating subpatterns, containing placeholders for specific word classes (subject, verb,\ldots). At a more superficial level, there are also often recurring semantical connections which more or less fix the word order: Let's just think of a formal context where every proper name has to be preceded by ``Mr.'' or ``Ms.''. One could think about expressing these relations, once retraced and extracted from a text, as logical rules, in particular as disjunctive normal forms, that is, as disjunctions of conjunctive clauses: ``either $A$ and $C$ happens, or $B$ and $D$''. This approach was suggested in \cite{FT1} and \cite{FT3}, where it was described as a sequence of minimum satisfiability problems. In \cite{FT2} the possible strategies are detailed for an adequate algorithm that can efficiently produce such logic formulas. Interesting about this whole process is that so found clauses can be arranged in a hierarchical decreasing order by their discriminating power; this allows to select them following some sort of survival of the fittest, given the importance assigned to the formulas modelling rarer word sequences present in training data. Returning to the ILP problem \eqref{ilp}, it is remarkably possible to translate conjunctive clauses into integer linear constraints.\\
The perhaps most used logical constraints could thus be: adjacency ($A$ should be immediately followed by $B$), precedence (if $A$ appears before the last token, $B$ should appear somewhere after that), state change (a punctuation mark $D$ should be preceded by $A$ and followed by $B$), begin-end position (if the sequence starts with $A$, it should end with $B$), presence and precedence (if $A$ appears, then $B$ shouldn't appear before~$A$).

\bigskip

The crucial point is that all the constraints mentioned before should, but must not take place \emph{a priori}. We want therefore to introduce a vector $\sigma$ whose dimension corresponds to the number of introduced constraints and whose entries are binary: 0 if the constraint is respected and 1 otherwise. If $H$ is the matrix modelling the logical constraints, we could represent their presence under the form \[H\cdot e-\sigma\leq0.\] It would take the following explicit forms for some of our previous mentioned examples:

\begin{itemize}
                               \item Adjacency:\[\sum_{y\in\tilde{\mathcal{Y}}}e_{t-1,yA}-e_{t,AB}-\sigma_c\leq0\quad\forall1\leq t\leq n-1\]
                               \item Precedence:\[\sum_{y\in\tilde{\mathcal{Y}}}e_{t-1,yA}-\sum_{z=1}^{n-t}\sum_{y\in\mathcal{Y}}e_{t+z,By}-\sigma_c\leq0\quad\forall1\leq t\leq n-1\]
                               \item Begin-end position: \[e_{0,y_pA}-e_{n,By_f}-\sigma_c\leq0\]
                             \end{itemize}

Here, $\sigma_c$ is the generic element of the vector $\sigma$ corresponding to a constraint.

\bigskip

The central problem now shifts from finding the most probable and shortest path on the graph, which we call $e^*$ and can find just using the Viterbi algorithm, to finding it applying the new constraints and at the same time minimizing the error rate. We define $c$ as the vector of the costs for breaking a constraint and then proceed to formulate the new problem:

\begin{equation}\begin{split}\label{ilp2} &\min W(e)=c^t\cdot \sigma\\ \text{subject to}\quad&M^t\cdot e\geq\tau Z(e^*)\\&\sum_{y\in\tilde{\mathcal{Y}}}e_{t-1,y\hat{y}}-\sum_{y\in\tilde{\mathcal{Y}}}e_{t,\hat{y}y}=0\quad\forall t\quad\text{t.c.}\quad1\leq t \leq n,\quad\forall \hat{y}\in\mathcal{Y}\\ &\sum_{y\in\mathcal{Y}}e_{0,y_py}=1,\quad\sum_{y\in\mathcal{Y}}e_{n,yy_f}=1\\&H\cdot e-\sigma\leq0.\end{split}\end{equation}

The first constraint is a lower bound which ensures us that the new solution we are looking for will be close enough to the optimal one of problem \eqref{ilp}; it is given in terms of percentage and $\tau$ is a real number lying in the interval $[0,1]$. The objective function $W$ represents the total cost for violating one or more constraints, 0 in the best case. In particular, every entry of $c$ might be determined as the logarithmic probability that its corresponding constraint will be violated. That is, given a clause $l$ representing the logical relationship between labels, the cost of violating all the constraints related to $l$ might be computed as \[c_{l}=\log P\left(\frac{|D(l)|}{|D(l)|+\overline{|D(l)|}}\right),\] where $D(l)$ denotes the set of true clauses and $\overline{D(l)}$ the set of clauses not satisfied in training data.

\bigskip

An alternative solution that allows us to include the constraints introduced by logical relations and errors, which appear to be not so elementary, in the proposed model is to relax the original problem \eqref{ilp} by mean of Lagrangian relaxation.

\subsubsection{Lagrangian Relaxation}

Sometimes, some constraints in an ILP problem are stumbling blocks that can render the solution too difficult to be obtained directly. Lagrangian relaxation permits to enunciate a dual problem whose solution is the same as the primal one, but easier to find with iterative methods (see as reference \cite{NF}, chapter 16).

\bigskip

In problem \eqref{ilp2}, although the focus is on minimizing the total errors' cost, we still have to find a path $e$ through the graph; this path has to be comparable to the optimal path of the non constrained version. However, we have added other constraints whose violation is penalized by some cost $c$. Now, in the precedent formulation, the vector $c$ has to be defined prior to solving the ILP problem, but such a prediction might not be obvious. Instead, we could revert to problem \eqref{ilp} and decide to directly penalize the objective function $Z$ with respect to the logical constraints. That is, we are taking a Lagrangian approach to this, where the primal problem is \eqref{ilp} augmented with the ``difficult'' constraints to be relaxed, expressed by the matrix $H$, defined as before. We obtain: \begin{equation}\begin{split}\label{ilp3} &\max Z(e)=M^t\cdot e\\ \text{subject to}\quad&\sum_{y\in\tilde{\mathcal{Y}}}e_{t-1,y\hat{y}}-\sum_{y\in\tilde{\mathcal{Y}}}e_{t,\hat{y}y}=0\quad\forall t\quad\text{t.c.}\quad1\leq t \leq n,\quad\forall \hat{y}\in\mathcal{Y}\\ &\sum_{y\in\mathcal{Y}}e_{0,y_py}=1,\quad\sum_{y\in\mathcal{Y}}e_{n,yy_f}=1\\&H\cdot e\leq0.\end{split}\end{equation}

We start relaxing it by defining a vector $\lambda$ of positive real numbers, the so-called Lagrangian multipliers. We then define the Lagrangian as \begin{equation}\begin{split}\label{lag}L(\lambda)&=\max\{M^t\cdot e-\lambda^tH\cdot e\}\\&=\max\{(M^t-\lambda^tH)\cdot e\},\end{split}\end{equation} where $e$ is defined as before. We can view $e$ as a variable lying in the space $E\subset\mathbb{R}^N$, for an appropriate $N$. It is easy to see that if $e'$ is an optimal solution for problem \eqref{ilp} (i.\ e. problem \eqref{ilp3} with no extra constraints), its corresponding value $z^*$ is always smaller than $L(\lambda)$, for every $\lambda$.\\
Now, we then define the Lagrangian dual as the problem to find \begin{equation}L^*=\min_{\lambda}L(\lambda).\end{equation} In fact, a fundamental theorem (see \cite{rillag}) assures us that $L^*$ coincides with the optimal solution of problem \eqref{ilp3}.

\bigskip

With this fact at hand, we can give a brief geometric and analytical description of $L(\lambda)$. Since the vector $e$ is made out of integers, there will be only a finite number $S$ of solutions $e_i$, $i=1,\ldots,S$, satisfying the request of $e$ being a path. Consequently, for every such $e_i$, if we fix it in the expression $M^t\cdot e-\lambda^tH\cdot e$, since this new expression is linear in the variable $\lambda$ it will describe a hyperplane in an appropriate space $\mathbb{R}^n$, whose dimension depends from the number of logical relations. As said before, there are $S$ of such hyperplanes, parametrized by the possible solutions $e_i$. Over every fixed point $\lambda$, as defined in equation \eqref{lag}, the value of $L(\lambda)$ is then given by the highest between these $S$ hyperplanes, since we are searching for the maximum. Then, it is easy to derive from this representation that $L(\lambda)$ is a convex, polytopic function with its minimum in $L^*$.\\This means that we could try to find $L^*$ starting from some point on the surface and going iteratively downstream following its gradient, i.\ e. the direction of the greatest rate of increase of the function, weren't the function not differentiable. We have therefore to recur to the subgradient.

\bigskip

The subgradient of a convex function $f(x)$, $x\in\mathbb{R}^n$, in the point $x^*$ is defined as a vector $s\in\mathbb{R}^n$ such that \[f(x^*)-f(x)\leq s^t(x^*-x).\] Then we can exploit a result (see \cite{rillag}) which concretizes the subgradient for our $L(\lambda)$; namely, if $e^*$ is a solution by which the value $L(\lambda^*)$ for a specific $\lambda^*$ is attained, we may write the subgradient of $L(\lambda)$ in $\lambda^*$ as \[-H\cdot e^*.\] Obviously, if the subgradient equals 0 we know that $\lambda^*$ is the point where the minimum is reached. So, the proposed strategy is to generate a succession of parameters $\lambda_0,\lambda_1,\ldots$ for which the corresponding succession of subgradients converges to 0. The algorithm could have the following form at every step $k$, taking $\lambda_0=0$ as its first value:
\begin{itemize}
  \item it computes the path $e_k$ which realizes $L(\lambda_k)$;
  \item it then computes the subgradient $s_k=-H\cdot e_k$. It stops if it's 0;
  \item it computes the step $\theta_k$;
  \item it computes the new parameter $\lambda_{k+1}=\lambda_k+\theta_k\frac{s_k}{||s_k||}$ and starts again.
\end{itemize}

The algorithm could be stopped after a fixed number $K$ of iterations if the estimation of $L(\lambda^*)$ hasn't improved.\\
The choice of the step is fundamental and has to be made carefully to let the algorithm converge quickly; furthermore, it has to satisfy the conditions \[\lim_{k\rightarrow\infty}\theta_k=0\quad\text{and}\quad\sum_{k=0}^{\infty}=+\infty.\] In our case, we used $\theta_k=\frac{1}{k+1}$.

\section{Experimental Results}
The proposed method has been tested with success on different benchmark data. The performance criteria used to evaluate and compare the performance of the proposed method are detailed in the following.

\subsection{Performance criteria}
The performance in terms of effectiveness has been measured by using
four well known evaluation metrics, i.\ e. F-Measure, Precision, Recall and Accuracy.
The F-Measure metric represents a combination of Precision and Recall typical of Information Retrieval.
Given a set of  labels $\mathcal{Y}$  we
compute the Precision and Recall for each label $y \in \mathcal{Y}$ as:
\begin{eqnarray}
Precision(y)=\frac{\textrm{\# of tokens successfully predicted as } y}{\textrm{\# of tokens predicted as } y}
\end{eqnarray}\
\begin{eqnarray}
Recall(y)=\frac{\textrm{\# of tokens successfully predicted as } y }{\textrm{\# of tokens effectively labelled as } y }
\end{eqnarray}\
The F-Measure for each class $y \in\mathcal{Y}$ is computed as the
harmonic mean of Precision and Recall:
\begin{eqnarray}
F(y)=\frac{2 \cdot Recall(y) \cdot Precision(y)}{Recall(y) +
Precision(y)}
\end{eqnarray}

The Accuracy measure can be summarized as follows:
\begin{eqnarray}
Accuracy=\sum_{y}^{\mathcal{|Y|}}\frac{\textrm{\# of tokens correctly labelled as
} y}{\textrm{total \ number\ of \ tokens}}
\end{eqnarray}
Considering that the data sets used in this experimental evaluation are composed of unbalanced samples, i.\ e. the class distribution of each label is not uniform, both micro- and macro-average have been computed  for Precision, Recall and F-Measure. Micro- and macro-measures differ in the computation of global performance:  macro-averaging gives equal weight to each label category (independently from category size), while micro-averaging considers the contribution of each label class according to  its dimension.

\subsection{Data set}
To evaluate the proposed inference method against the traditional one, we have performed experiments on the data set \emph{Cora}. The \emph{Cora} citation benchmark is composed of 500 citations of research papers annotated with 13 different labels: \emph{Title, Author, Book Title, Date, Journal, Volume, Technology, Institution, Pages, Editor, Location, Notes}.  The benchmark has been split for training and  testing the models: 350 instances have been used as training set, while the remaining 150 instances as testing.
\begin{table}[htbp]
\begin{center}\tabcolsep=6.5mm
{\renewcommand\arraystretch{1.2}
\begin{tabular}{l|c|c}
 & \multicolumn{2}{c}{\textbf{Cora}} \\
 &      \textbf{Train} &       \textbf{Test} \\
\hline
Author &    1948        &        801 \\
\hline
 Title &       2585     &       1103  \\
\hline
Publisher &    226        &         61 \\
\hline
Booktitle &     1414       &        477 \\
\hline
Data &    452        &        187  \\
\hline
Journal &402&        219  \\
\hline
Volume &     216       &        104  \\
\hline
Technology &    173        &         57 \\
\hline
Insitution &    236        &         71  \\
\hline
Pages &   500         &        208  \\
\hline
Editor &    151        &         74 \\
\hline
Location &    208        &         95      \\
\hline
Note &     116       &         17  \\
\hline\hline
  \textbf{Tot.} &     \textbf{8627}       &    \textbf{3474}     \\
\hline
\end {tabular}\\}
\caption*{\emph{Cora} label distribution}
\end{center}
\end{table}

The results obtained on \emph{Cora} are as follows, respectively for the basic, non constrained problem solved by Viterbi algorithm and the proposed inference method with and without Lagrangian relaxation.

\scriptsize
\[                                                                                                                                                           \begin{tabularx}{\textwidth}{lXXcXXcX}                                                                                                                      & \multicolumn{3}{c}{Macro-average} & \multicolumn{3}{c}{Micro-average} &  \\                                                                              \hline                                                                                                                                                   & Precision & Recall & F-measure & Precision & Recall & F-measure & Accuracy \\                                                                            \midrule                                                                                                                                                    Viterbi & \textbf{85.49} & 72.02 & 76.36 & 87.27 & 87.1 & 86.65 & 87.13\\                                                                                   Constraints & 81.34 & \textbf{78.95} & \textbf{79.93} & 90.53 & 90.32 & 90.27 & 90.32\\                                                                     Lagr.\ Relaxation & 81.58 & 77.82 & 79.14 & \textbf{90.87} & \textbf{90.92} & \textbf{90.66} & \textbf{\textbf{90.36}} \\                                   \hline                                                                                                                                                  \end{tabularx}                                                                                                                                              \]                                                                                                                                                          \normalsize

\section{Conclusions}

Starting from the state-of-the-art method of Conditional Random Fields used in the tasks of Named Entity Recognition, in this paper we have shown how it is
possible to translate the CRF approach, in its most simple but useful case when we are dealing with linear chains, to a problem of Integer Linear
Programming (\cite{crf}). This enables us to solve it with the Viterbi algorithm. However, to model even long-distance relationships in the
text, which could otherwise be lost, we decided to add new linear constraints to the original problem. The fact that such constraints are added to the ILP
formulation means that the relationships are modelled during the inference phase, enabling us to face a much lesser computational load than if they had
been added to the original CRF model itself. Thanks to the work found in \cite{FT1}, \cite{FT2} and \cite{FT3} we know an algorithm capable of extracting
relationships of the aforementioned type under the form of logical formulas, which can in turn be easily converted to linear equations and arranged in
order of importance. The next step is to consider the possibility that some constraints won't be satisfied, thus improving the
flexibility of the prediction, which could have the risk to incur into unfeasible options. A measure of the total error committed is therefore introduced,
and a new ILP problem is formulated in which the objective function tries to minimize it. This happens at the same time requiring that new solutions do not
strive too far away from the one of the original problem without logical constraints. The newly found constraints are however difficult to calculate
directly, so that numerical methods are best used. The idea is then to apply Lagrangian relaxation to the original problem augmented with the logical
relations, so as to penalize the objective function for errors, without making their cost explicit. It is proved by some fundamental theorems (see \cite{NF} or \cite{rillag}) that a dual of the problem is easily constructed and that its minimum, corresponding to the optimal solution of the primal problem, can be determined through an iterative algorithm. Nonetheless, the challenge is left to aptly attune this algorithm and to understand how quickly it converges and how it is possible to practically implement it in the specific cases presented here.

\end{document}